\documentclass[letterpaper, 10 pt, conference]{ieeeconf}
\IEEEoverridecommandlockouts
\usepackage{graphics} 
\usepackage{epsfig} 
\usepackage{mathptmx} 
\usepackage{times} 
\usepackage{amsmath} 
\usepackage{amssymb}  
\usepackage{url}
\usepackage{soul}
 \usepackage{subfigure}
\usepackage{multicol,tabularx,capt-of}
\usepackage{multirow}

\usepackage{color}

\title{\LARGE \bf
On Simple Reactive Neural Networks for Behaviour-Based Reinforcement Learning
}

\author{Ameya Pore$^{1}$ and Gerardo Aragon-Camarasa$^{1}$
\thanks{$^{1}$ Computer Vision and Autonomous group, School of Computing Science, University of Glasgow, G12 8QQ, UK. Email: {\tt\small ameya.pore@students.iiserpune.ac.in, gerardo.aragoncamarasa@glasgow.ac.uk}}%
}

\begin{document}

\maketitle
\thispagestyle{empty}
\pagestyle{empty}


\begin{abstract}
We present a behaviour-based reinforcement learning approach, inspired by Brook's subsumption architecture, in which simple fully connected networks are trained as reactive behaviours. Our working assumption is that a pick and place robotic task can be simplified by leveraging domain knowledge of a robotics developer to decompose and train reactive behaviours; namely, \textit{approach}, \textit{grasp}, and \textit{retract}. Then the robot autonomously learns how to combine reactive behaviours via an Actor-Critic architecture. We use an Actor-Critic policy to determine the activation and inhibition mechanisms of the reactive behaviours in a particular temporal sequence. We validate our approach in a simulated robot environment where the task is about picking a block and taking it to a target position while orienting the gripper from a top grasp. The latter represents an extra degree-of-freedom of which current end-to-end reinforcement learning approaches fail to generalise. Our findings suggest that robotic learning can be more effective if each behaviour is learnt in isolation and then combined them to accomplish the task. That is, our approach learns the pick and place task in 8,000 episodes, which represents a drastic reduction in the number of training episodes required by an end-to-end approach (~95,000 episodes) and existing state-of-the-art algorithms.
\end{abstract}

\section{INTRODUCTION\label{sec:intro}}

Robots excel at using pre-programmed routines to perform repetitive tasks. A significant barrier in their universal adoption beyond an enclosed environment is their fragility and lack of robustness in complex environments. To tackle these issues, recent advances in deep learning and deep Reinforcement Learning (RL) have enabled the development of robotic solutions for complex and diverse scenarios that have been intractable using classic traditional control approaches. Examples include decision making for solving games \cite{mnih2013playing, silver2016mastering}, and continuous control tasks such as locomotion skills, dexterous manipulation and grasping \cite{haarnoja2018learning, gu2017deep, haarnoja2018composable, vid2param}. However, a limitation to the widespread adoption of RL algorithms in robotics is that RL approaches dramatically overfits the idiosyncrasies of training environments \cite{packer2018assessing, zhang2018dissection}.

\begin{figure}[t]
\centering
    \includegraphics[width=0.45\textwidth]{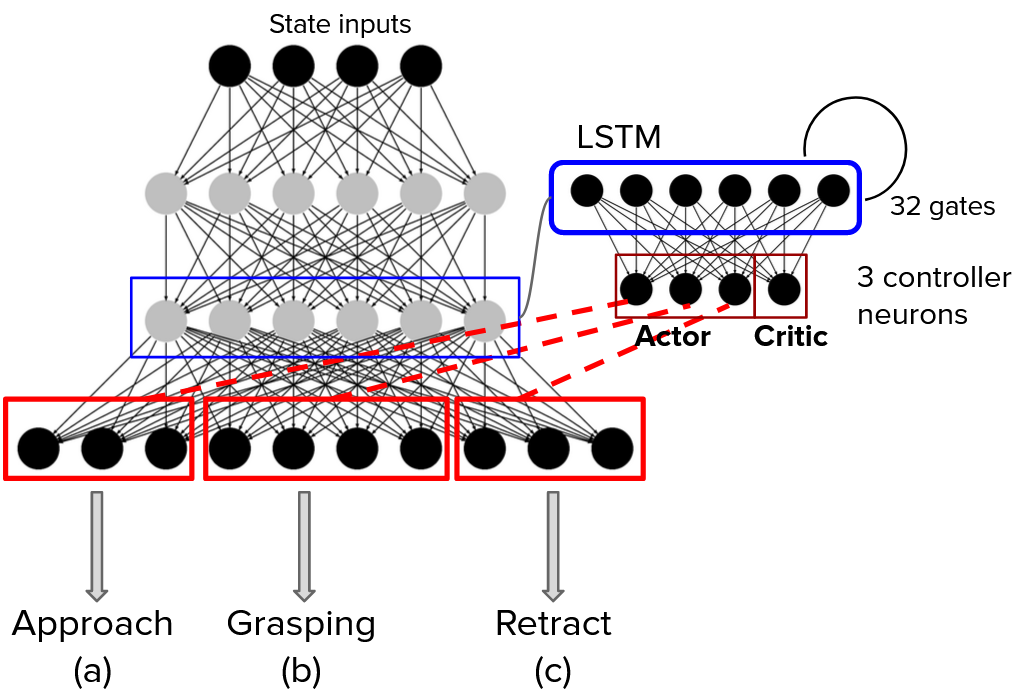}
    \caption{Schematic of the modular behaviour-based reinforcement learning architecture. The goal of picking an object is subdivided into simpler behaviours that are trained specifically for movements in x, y, z and $\theta$ of the end-effector. These behaviours are activated and inhibited by a reactive network parametrised by an actor-critic policy. See Section \ref{sec:materials} and Fig. \ref{fig:fig2} for details.}
    \label{fig:fig1}
\end{figure}

In this paper, we depart from an end-to-end RL approach, and we investigate whether it is possible to train a robot to pick up a block using manually predefined behaviours that are then choreographed using RL. Specifically, we propose a modular behaviour-based reinforcement learning architecture (see Fig. \ref{fig:fig1}) that is inspired by the subsumption architecture \cite{brooks1991intelligence}. That is, we start by training neural networks with a known solution, leveraging on domain knowledge of a robotics developer. We, therefore, guide a robot to learn specific low-level behaviours. Once these low-level behaviours are acquired, an RL algorithm explores how to choreograph behaviours in different temporal combinations; effectively learning to subsume behaviours that are not relevant to the current state input. We validate our approach in a simulated environment -- MuJoCo simulator using the \textit{FetchPickandPlace} environment \cite{brockman2016openai, plappert2018multi}. We must note that we go beyond the basic environment structure and allow the block to spawn in random positions and orientations to include an additional degree of freedom (gripper rotation) while grasping the block (Fig. \ref{fig:fig1}(b)). Our contributions are therefore twofold: (i) an architecture that learns isolated modular behaviours that which (ii) drastically reduces the number of steps required to train a robot while performing a pick and place task, i.e. picking and placing a block in a simulated environment. The source code for this paper is available at \url{https://github.com/cvas-ug/simple-reactive-nn}. A video summary of this paper can be found at \url{https://youtu.be/z7kUW9yyka0}.

\section{RELATED WORK}

Brooks proposed the Subsumption Architecture (SA) \cite{brooks1991intelligence, brooks1990elephants} to mimic the evolutionary path of intelligence. This architecture consists of designing simple behaviours to achieve robust and complex behaviours in robots by layering them in terms of complexity and execution time. Specifically, the SA connects perception to action for robot control systems and coordinates defined behaviours. In SA, complex behaviours subsume a set of simple behaviours, and a task is accomplished by activating the appropriate behaviour given an input state. Hierarchical Reinforcement Learning (HRL) \cite{nachum2018data,levy2018learning,kober2013reinforcement} resembles the SA in the sense that a complex task is automatically decomposed into sub-task sequences, that are themselves built by machine-defined simple actions. HRL learns and operates at different levels of temporal abstraction by using multiple layers of policies that are trained to perform decision-making and control at the successively higher level of behavioural and temporal abstractions. The lowest-level policy of the hierarchy (subordinate actions) applies actions to the environment, whereas the higher-level policies are trained over a longer time scale.

Current HRL approaches include the work by Nachum et al. \cite{nachum2018data}, where the authors propose HIRO, a 2-level HRL approach that can learn off-policies. While Levy et al. .\cite{levy2018learning} have built a hindsight experience with similar hierarchical architecture as HIRO to increase the sample efficiency in sparse reward conditions. The key distinction between these HRL approaches and our approach is that in HRL, multiple policies are learnt in parallel and end-to-end, whereas we attempt to learn them incrementally. Concretely, HRL algorithms autonomously decide how to segment the main task into sub-tasks. This segmentation is task-specific, and the decomposed sub-task, once trained, would hardly be able to generalise to a different high-level task (Section \ref{sec:intro}). Moreover, learning multiple behaviours end-to-end leads to the curse of dimensionality as the task becomes temporally elongated \cite{kober2013reinforcement}. 

Our work is closely related to the architecture proposed by \textit{Konidaris et al.} \cite{konidaris2005architecture} in behaviour-based reinforcement learning, where a topological map is learnt to create task-relevant state spaces, and layered reinforcement learning takes place over this map.  Multiple learning models, multiple control processes, and a complex environment result in complex learning behaviours. However, a significant drawback of this approach is that it is not feasible to build topological maps in many situations. We have addressed this by using neural networks that learn feature representations from raw sensory data. Similarly, the work by Frans et al. \cite{frans2017meta} presents an end-to-end method to use shared policy primitives, within a distribution of tasks, and are switched between by task-specific policies to execute over a large number of timesteps. A master policy is learnt, and this policy selects a sub-policy to be active. In this paper, we use shared parameters for state feature representation to train a policy that selects the underlying trained primitive policies, and the primitive policies are incrementally trained.

From the above, we can observe that state of the art approaches adopts an end-to-end strategy to optimise and learn tasks and sub-tasks. However, humans tend to learn simple behaviours first in order to compose complex behaviours \cite{bengio2009curriculum}. For example, while learning tennis, we start by learning \textit{basic behaviours} separately such as bouncing the ball, hitting the ball, serve, tp name but a few. In contrast, an end-to-end approach would attempt to optimise all possible behaviours mat the same time. That is, an artificially intelligent agent requires millions of trials using sophisticated model-free RL algorithms to complete simple tasks on simulations and games, whereas humans learn behaviours in 50-100 attempts \cite{dubey2018investigating}. RL agents start solving each problem \textit{tabula rasa} 
with no human expert used as part of the training, whereas we come in with a wealth of prior knowledge about the world, from physics to semantics to affordances. In this paper, we propose that a robot learns basic and simple behaviours. After building a set of these behaviours, a robot can then learn how to choreograph these autonomously using reinforcement learning.

\section{MATERIALS \& METHODS\label{sec:materials}}

\begin{figure*}[t]
    \centering
    \includegraphics[width=\textwidth]{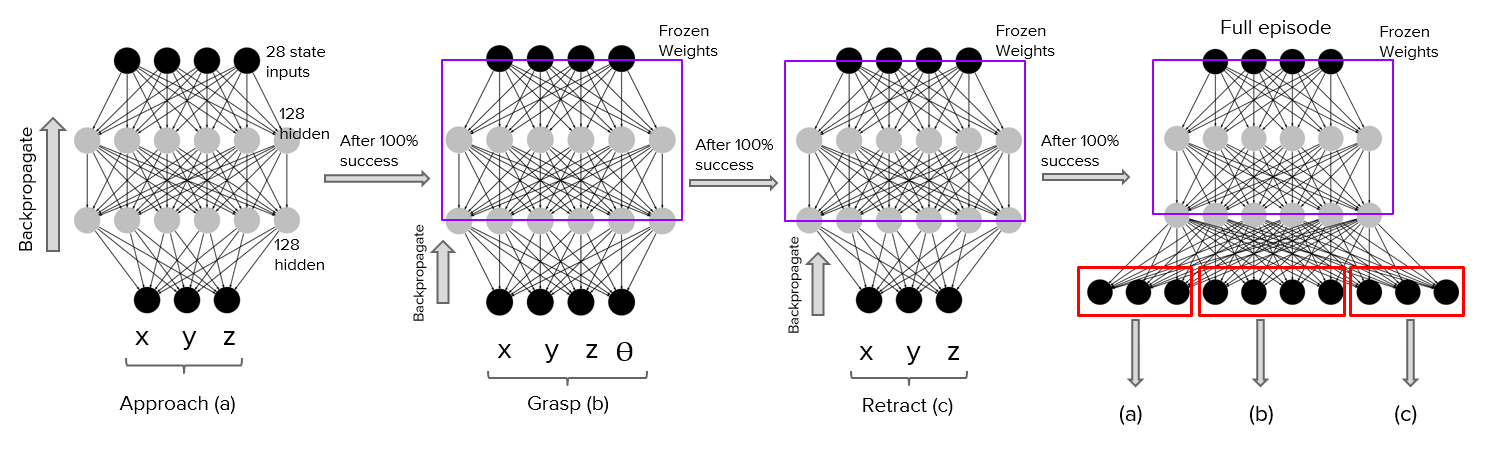}
    \caption{The network takes as input a 28 state vector and outputs the $x$, $y$, $z$ movement of the end-effector. The modules of \textit{approach} (a), \textit{grasp} (b) and \textit{retract} (c) are trained separately using behaviour cloning and combined to accomplish the task. First, the state vector is given as an input to the feature extraction layer. The extracted features are relayed on to the reactive layers.}
    \label{fig:fig2}
\end{figure*}

\subsection{Rationale\label{sec:rationale}}

Our working assumption is that human knowledge can simplify the task of a robot solving a pick and place task. Hence, we manually break up tasks into simple behaviours and enable the robot to sequence them according to the state encountered. The goal of the robot is to pick and take the block to a target location. We, therefore, decompose the main pick and place task into \textit{approaching}, \textit{grasping a block}, and \textit{retracting to a target point} simple behaviours (Fig. \ref{fig:fig2}). These low-level behaviours start with general and primitive abilities that are controlled, overridden, or subsumed by specific goal-directed behaviours. We, therefore, adopt the SA architecture as the means of defining behaviours to enable an incremental and sequential, bottom-up operation of the system.

The overall behaviour of the robot is thus a consequence of the responses within the environment. Behaviours rely on the state of the world without maintaining a global internal representation \cite{brooks1991intelligence}. We use a supervised learning approach to learn from demonstrations instead of learning new policies from scratch (Section \ref{sec:lowlevelbx} \cite{nair2018overcoming, bojarski2016end}. That is, if the robot is approaching the block, we explicitly teach it to move in the $x$, $y$ and $z$ Cartesian space. We further repeat this for other behaviours. Once, these low-level behaviours are learnt, we train an actor-critic RL architecture (Section \ref{sec:sequencer}). The RL architecture determines the activation and inhibition mechanisms of low-level behaviours in a particular temporal sequence that ultimately give rise to the high-level behaviour of picking and placing a block.

\subsection{Low-level Behaviours\label{sec:lowlevelbx}}

Under the Subsumption Architecture, behaviours do not own memory and are decomposed in layers, each with a predefined goal \cite{brooks1991intelligence}. We, therefore, train a neural network to learn a specific behaviour using demonstrations of low-level behaviour. For this, we use behaviour cloning (BC), which learns a policy through supervised learning in order to mimic the demonstrations \cite{nair2018overcoming, bojarski2016end, nakanishi2004learning}. Expert demonstrations of successful behaviours are used to train a network which learns to imitate the expert providing these successful trajectories \cite{hester2018deep}.

We thus use a loss function computed on the demonstration examples as follows:
\begin{align}
L_{BC} = ||\pi (s_i |\theta_\pi ) - a_i ||^2
\end{align}
where $a_i$ refers to the intended output of the behaviour. $\pi(s_i)$ refers to the action predicted by the robot at state $s_i$ under the policy parametrised by $\theta(\pi)$. We reparameterise the policy such that a sample from $\pi_\theta(\cdot|s_i)$ is drawn by computing a deterministic function of the state, policy parameters and independent noise. We use a neural network transformation $a_i = f(\phi_i; s_i)$, where $\phi_i$ is an input noise vector, sampled from a fixed distribution, such as a spherical Gaussian,
\begin{align}
a_i = \tanh{(\mu_\theta(s_i)+ \sigma_\theta(s_i)\phi_i)},    
\phi_i \sim N(0,1)     
\end{align}

The raw input of the kinematic coordinates and velocity of block and gripper are fed to a feature extraction network consisting of two fully connected layers with 128 neurons each. The output is mapped to six neurons in the last layer that determines the mean and standard deviation of a Gaussian distribution from which the movement values are sampled for the end-effector in $x$, $y$ and $z$ for approach(a), grasping (b) and retract (c). For training (Fig. \ref{fig:fig2}), we start with the approach (a) behaviour module (Fig. \ref{fig:fig2}). That is, the output of the network is used to control the end-effector while it is approaching the block (i.e $\mid d_{eff}- d_{block}\mid <error$)\footnote{$d_{eff}$ refers to position of end-effector, $d_{block}$ refers to the position of the block, and $error$ is manually set to $0.01$ for training (a) and (c), and $0.005$ for training (b)} and we use hand-engineered solutions for grasping (b) and retract (c).

The BC loss function is backpropagated after each step while approaching. Once the success rate reaches its maximum, training is stopped, and we save the weights. Then, we train the grasp (b) behaviour module. For this, the weights of the feature extraction (the first two layers) are frozen, and a similar training process using BC is carried out where the output of the network is used to control the end-effector while it is grasping the block. Finally, for training the retract module, we froze the weights of (a) and (b) and a similar training strategy is carried out until it reaches maximum success rate for (c); in this case, we minimise the distance between the end-effector and the target point (i.e. $\mid d_{target}- d_{block}\mid <0.01$).

\begin{table*}[th]
    \centering
    \caption{Experimental strategies for training reactive behaviours before interfacing them with the high-level choreographer.}
    \label{tab:experiments}
    \begin{tabular}{|m{5mm}|m{15mm}|m{140mm}|}
        \hline
        \textbf{\#} & \textbf{Name} & \multicolumn{1}{c|}{\textbf{Training Strategy}} \\ \hline
        1 & Sequential & The approach behaviour (a) with a hand-engineered solution for grasping (b) and retract (c) behaviours. Once (a) is trained (i.e. reaches a desired performance level), (b) is trained using the output from the network (a) and a hand-engineered solution for (c). Similarly, once (b) is learnt, we train on (c), using outputs from the (a) and (b) networks. \\ \hline
        
        2 & Sequential + Freezing & We start by training (a), with a hand-engineered solution for (b) and (c). After (a) is trained, we freeze the layers that extract features from the raw input. We then train (b), using the frozen weights for feature extraction and the output of (a), and a hand-engineered solution for (c). A similar approach is applied for scaling up and training on (c). \\ \hline
        
        3 & Separate & We start by training (a) with a hand-engineered solution for (b) and (c). Once, (a) is trained, we do not use the output of the trained network (a). For training (b), we use a hand-engineered solution for (a) and (c). Similarly, training is carried out for (c). Once, each module is trained separately; we combine all the behaviours to accomplish the task. \\ \hline
        
        4 & Separate + Freezing & We start by training (a) with a hand-engineered solution for (b) and (c). After (a) is trained, we freeze the layers that extract features from the raw input. For training (b), the output from the network (a) is not used. Instead, we use a hand-engineered solution for (a) and (c) with frozen weights of the feature extraction layer. Similarly, training is carried out for (c). Once, each module is trained separately; we combine all the behaviours to accomplish the task. \\ \hline
        
        5 & End-to-end & In this case, we do not decompose simple behaviours. That is, state inputs are directly mapped to actions training using behaviour cloning for all time steps in the episodes. The action space consists of only the $x$, $y$ and $z$ coordinates of the end-effector without considering the end-effector orientation, $\theta$. \\ \hline
    \end{tabular}
\end{table*}

\subsection{High-level Choreographer\label{sec:sequencer}}

As stated in Section \ref{sec:rationale}, the high-level sequencer learns a policy that choreographs a set of behaviours in order to solve a robotic pick and place task. We consider the standard Markov decision process framework for picking optimal behaviours to maximise rewards over discrete timesteps in an environment $E$ \cite{sutton1998introduction}. At every timestep $t$, the robot is in a state $s$, executes a behaviour $u_t$, receives a reward $r_t$, and $E$ evolves to state $s_{t+1}$. Lets now denote the return by $R_t = \sum_{i=t}^{T} {\gamma}^{(i-t)} r_i$, where $T$ is the horizon that the robot optimises over, and ${\gamma}$ is a discount factor for future rewards. The robot’s objective is to maximise the expected return from the start distribution,
\begin{align}
J = E_{r_t,s_t{\sim}E,u_t{\sim}\pi}[R_0].
\end{align}

For the high-level choreographer, the extracted features from the block's position (Section \ref{sec:lowlevelbx}) are passed through an LSTM layer with $32$ units. Two separate fully connected layers are used to predict the value function and the activation/inhibition from the LSTM feature representation -- see Fig. \ref{fig:fig1}. The aim of using a recurrent layer is that the agent should have a local memory of the amount of task it has accomplished.
In this paper, we adopted and tailored the Asynchronous Actor-Critic architecture (A3C) \cite{mnih2016asynchronous} to serve as our higher level choreographer. This architecture learns to sequence the low-level behaviours described in Section \ref{sec:lowlevelbx}, and consists of a neural network called the actor that predicts actions, and a network called critic that learns to predict the value of a state-behaviour pair by optimising the Q-function. We have used generalised advantage estimation to optimise the actor-critic model \cite{schulman2015high}.

\section{EXPERIMENTS\label{sec:experiments}}

\begin{figure}[t]
\centering
    \includegraphics[width=0.45\textwidth]{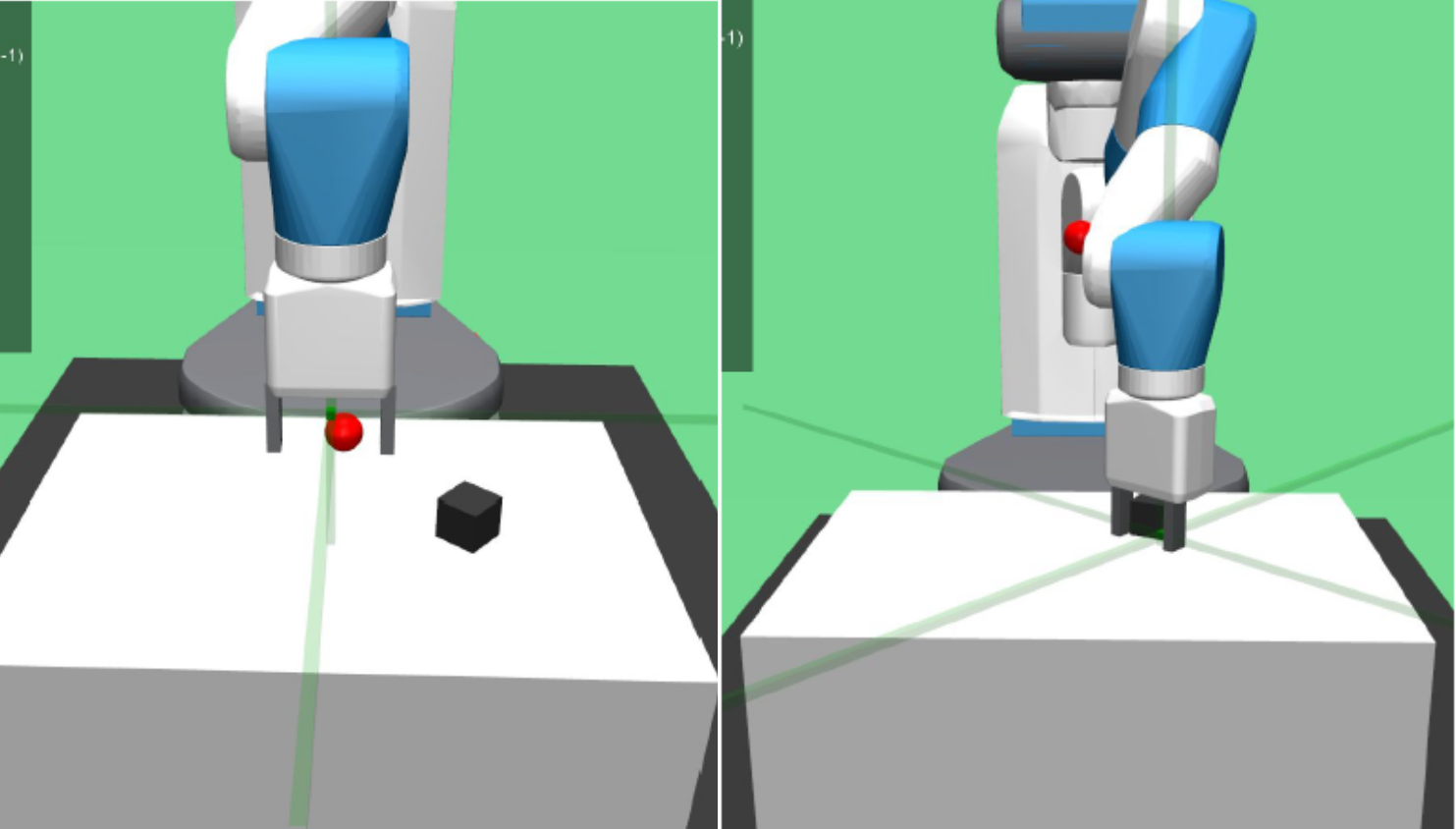}
    \caption{OpenAI \textit{FetchPickandPlace} environment as used in this paper.}
    \label{fig:simulation}
\end{figure}

In our experiments, we use a simulated environment based on the OpenAI \textit{FetchPickandPlace} environment \cite{brockman2016openai, plappert2018multi}. This environment is used as a benchmark for testing algorithms for continuous control tasks such as robotic manipulation and grasping. The goal in \textit{FetchPickandPlace} environment is to grasp a randomly positioned block and lift it to a target position. The environment provides kinematic values of position, velocity and orientation of the block and the gripper. In most studies related to RL algorithms controlling the gripper in Fetch, the orientation of the block is fixed in order to reduce the task complexity. We, however, activate the orientation of the gripper to grasp different block orientations, as shown in Fig. \ref{fig:simulation}.

\begin{figure*}[t]
    \centering
    \begin{subfigure}
        \centering
        \includegraphics[width=0.49\textwidth]{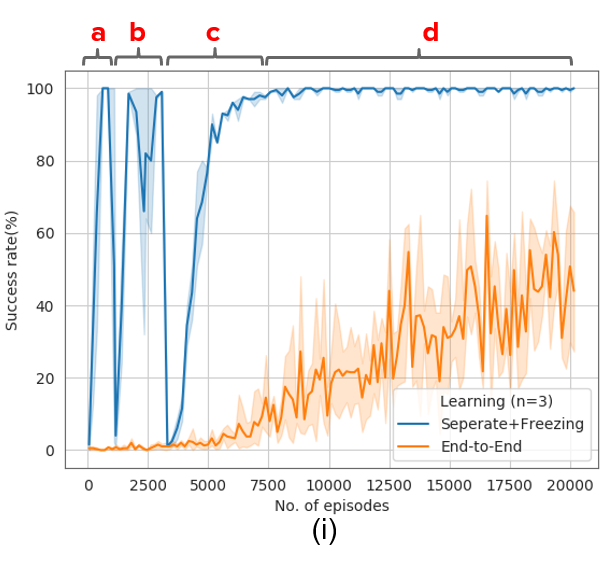}
    \end{subfigure}
    \begin{subfigure}
        \centering\includegraphics[width=0.49\textwidth]{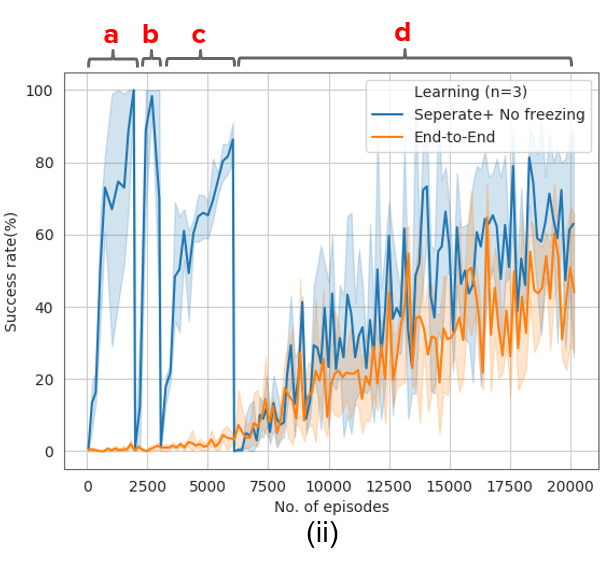}
    \end{subfigure}
    \begin{subfigure}
        \centering\includegraphics[width=0.49\textwidth]{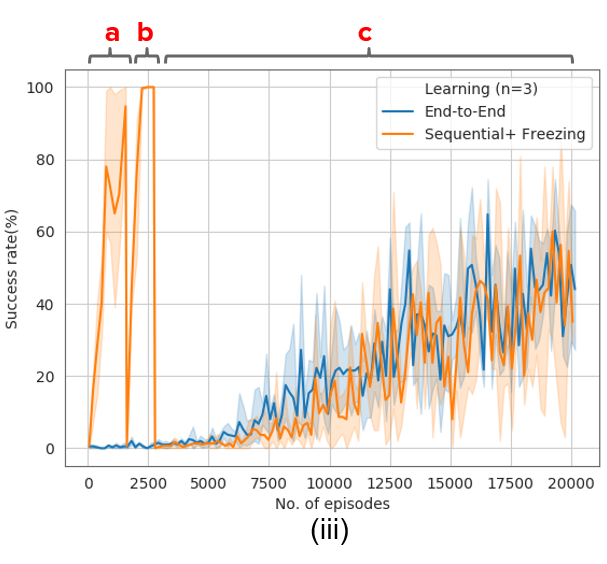}
    \end{subfigure}
    \begin{subfigure}
        \centering\includegraphics[width=0.49\textwidth]{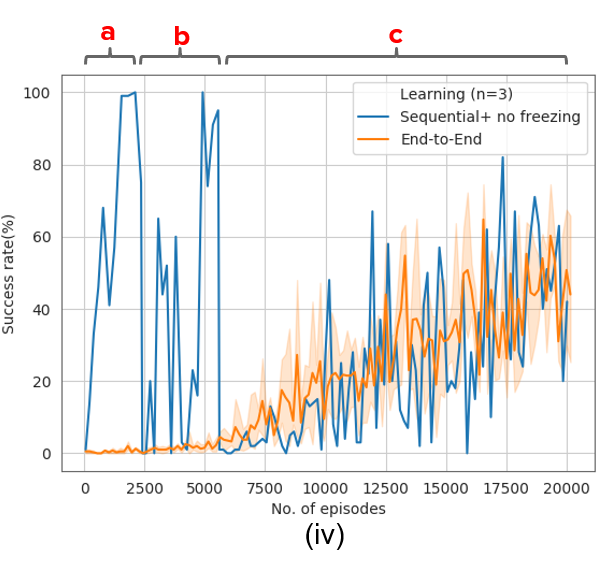}
    \end{subfigure}
    \caption{Comparison between end-to-end and the 4 proposed training strategies; i) Separate + Freezing (Yellow) and end-to-end (Blue); ii) Separate (Blue) and end-to-end (yellow); iii) Sequential + Freezing (Blue) and end-to-end (yellow); and iv) Sequential (blue) and end-to-end (yellow). Each training strategy is run independently for three runs (Learning $n=3$) with different seeds.}
    \label{fig:fig4}
\end{figure*}

In order to investigate which training strategy is the most optimal within our behaviour-based approach, we design five different training strategies, as described in Table \ref{tab:experiments}. We must note that the sequence of the behaviours is controlled manually in these training strategies, and behaviours are trained following the approach in Section \ref{sec:lowlevelbx}. For the end-to-end training strategy, we use a similar network structure for the feature extraction network as described in Section \ref{sec:lowlevelbx}. The output is then mapped to 6 neurons in the last layer that determines the mean and standard deviation of a Gaussian distribution from which the movement values are sampled for the end-effector in $x$, $y$ and $z$.

We then select the best performing training strategy and train the high-level choreographer, as described in Section \ref{sec:sequencer}. After training the high-level choreographer, we evaluate two scenarios: 

\begin{enumerate}
\item Sparse reward condition: The robot receives a reward after it picks the block to a target position. 
\item Dense reward condition: The robot receives a reward after each successful completion of a behaviour. For example: if the robot selects approach at the start of the episode, it receives a positive reward, on the contrary, if the robot selects other behaviours, it receives negative rewards. 
\end{enumerate}

\section{RESULTS\label{sec:results}}

Results are shown in Fig. \ref{fig:fig4}. The first peak in Fig. \ref{fig:fig4}(i-iv) denotes the completion of training network (a) (approach) when the success rate reaches 100\%, the second peak denotes the completion of training network (b) (grasping), and the third peak denotes the completion for the network (c) (retract). After each behaviour is trained, the success rate drops to zero since learning the new behaviour starts from scratch. Step (d) denotes the manual combination of (a), (b) and (c) to complete the task. In the end-to-end case (i.e. training strategy 5 in Table \ref{tab:experiments}), actions are optimised for the entire pick-and-place task. 

We can observe that it takes a similar number of training episodes for the end-to-end, \textit{Sequential + Freezing} (Fig. \ref{fig:fig4}-iii) and \textit{Sequential + No Freezing} (Fig. \ref{fig:fig4}-iv) approaches. \textit{Separate + No Freezing} (Fig. \ref{fig:fig4}-ii) shows a slight increase in the success rate, but its trend is close to the end-to-end approach. However, \textit{Separate + Freezing} (Fig. \ref{fig:fig4}-i) shows a drastic increase in success rate. In this case, the robot learns each skill separately with freezing layers after initial training, and our behaviour-based approach can reach 100\% success rate in 6,000 episodes of training. This result suggests that \textit{learning can be more effective if each skill is learnt in isolation and then combined in order to learn the high-level task of pick and place}. Secondly, once the feature extraction network learns the latent feature space, \textit{freezing the knowledge shows more learning potential for subsequent behaviours} (e.g. Fig. \ref{fig:fig4}-i). That is, once the robot knows how to perceive the state inputs, there is no advantage to learn the feature extraction layers again.

\begin{figure}[t]
    \centering
    \includegraphics[width=0.4\textwidth]{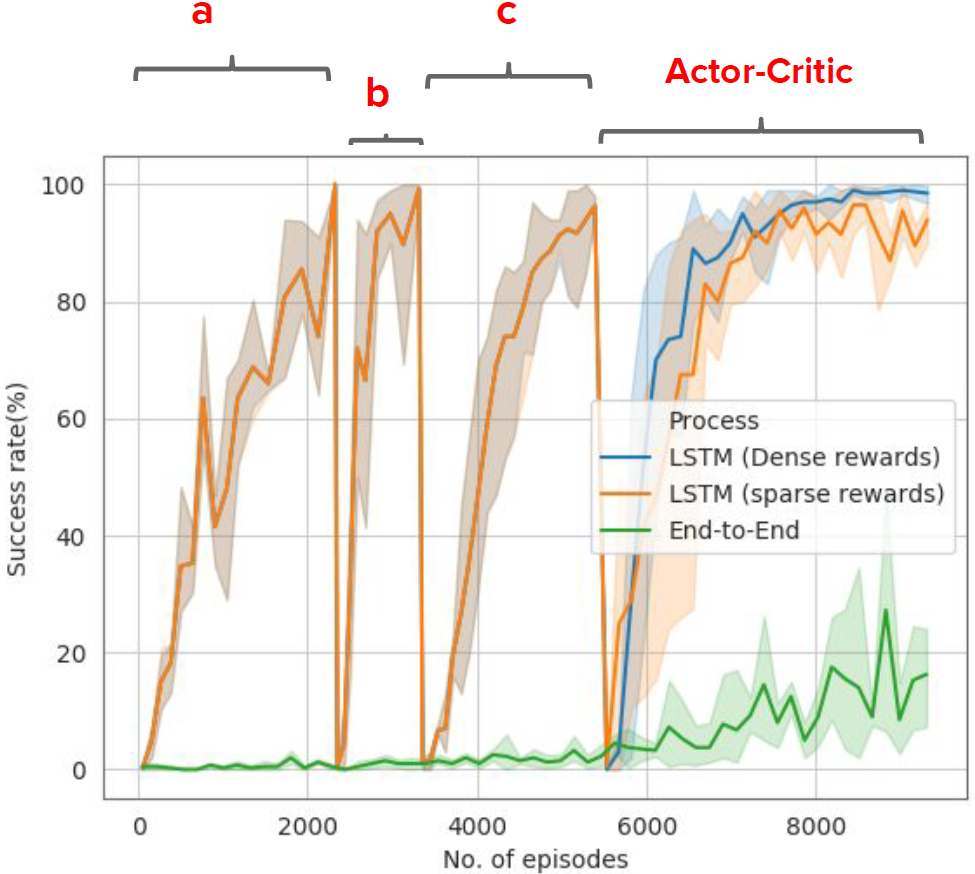}
    \caption{Comparison between the end-to-end approach (green) with the complete behaviour-based RL architecture (i.e. \textit{Separate + Freezing} combined with the high-level choreographer -- blue and orange).}
    \label{fig:fig5}
\end{figure}

The best performing strategy is the \textit{Separate + Freezing} (Fig. \ref{fig:fig4}-i), and we, therefore, use this strategy for the higher-level choreographer to learn how to sequence these behaviours. The choreographer uses the features from the feature extractor and selects the low-level actions based on the feedback of the rewards provided, as described in Section \ref{sec:sequencer}. This represents a simple RL setting where an agent has to decide the actions to maximise the total cumulative reward. Hence, we start training the high-level choreographer after 6,000 episodes (see Fig. \ref{fig:fig5}). For this, we evaluate two RL agents that receive rewards on different time-scales, namely dense and sparse reward settings (Section \ref{sec:experiments}. As expected, the RL agent that receives dense feedback can accomplish the task faster; however, the RL agent that receives sparse rewards performs close to the dense reward agent. One reason for this is that the pick and place task considered in this paper is short-time horizon and requires only three behaviours. We speculate that the robot with a dense reward condition would outperform the sparse reward setting agent in a log-term horizon task. This is left for future work as described in Section \ref{sec:conclusion}.

By further inspecting Fig. \ref{fig:fig5}, we can observe that the \textit{Separate + Freezing} combined with the high-level choreographer achieves 100\% success accuracy in approx. 8,000 episodes, effectively using only 2,000 episodes for training the high-level choreographer. This result represents a drastic reduction in the number of training episodes required by an end-to-end approach and the existing state-of-the-art RL algorithms such as Deep Deterministic Policy Gradients(DDPG) + Hindsight experience reply(HER) which takes 95,000 episodes to learn the grasping task \cite{plappert2018multi}. For our end-to-end approach, the robot is able to reach a maximum of 60\% success rate after 100,000 episodes. We speculate that the reason for this is due to the network shallowness to learn multiple behaviours at once, and backpropagation gets stuck in a local minimum. However, the latter is outside the scope of this paper.

Our results suggest that training can be more effective if each module is trained separately with other modules already trained; similar to what has been found in curriculum learning \cite{bengio2009curriculum}. Also, freezing the feature extraction layer after initial training and using this layer for training other modules, shows a considerable reduction in training time. This indicates that training of complex learning systems should be accomplished in a structured fashion, i.e. training simple modules first and independent of the rest of the network. It aligns with the paradigm of bottom-up learning approach where complex behaviours could arise by generating and combining simple ones \cite{gomes2018approach} of which motivates the adoption of Brook's Subsumption Architecture \cite{brooks1991intelligence} in this work.

\section{CONCLUSIONS \& FUTURE WORK\label{sec:conclusion}}

In this paper, we have used a bottom-up approach for training these
modular behaviours, i.e. Subsumption Architecture \cite{brooks1991intelligence}. We have demonstrated, in simulation, that long-time tasks can be decomposed and can be learnt independently. The latter can give rise to behaviours that could accomplish a variety of tasks. We train all the simple behaviours independently and then combine them sequentially to complete the task. We argue that these decomposed behaviours once trained could be used to accomplish different tasks and are task agnostic. Further, the proposed behaviour-based RL architecture is a simple feed forward neural network that maps the positional coordinates and kinematic state input to low-level actions and high-level behaviours via a learned and distributed internal representation.

From the results, we can state that our approach can learn to pick up a block in approximately 8,000 episodes, as opposed to an end-to-end learning approach and state-of-the-art RL approaches that take 95,000 episodes \cite{plappert2018multi} on simulation. The latter suggests that finding solutions using a model-free approach is data inefficient and requires several trial and error iterations for an RL agent to solve a task which is impractical in robotics. Our results also suggest that by tapping into human knowledge to decompose simple behaviours and separately learning these simple behaviours show a drastic reduction in training time. For future work, we will deploy our approach in a real robotic task to investigate how knowledge acquired by in simulation can be generalised to other types of objects. For this, we will use deep learning solutions for object recognition and for estimating the pose of objects, e.g. \cite{Billings19}, and investigate the use of continuous perception \cite{MARTINEZ2019220} to maintain temporal consistency during task execution.

\section*{ACKNOWLEDGEMENT}
We thank Paul Siebert, Ali Al-Qallaf, Piotr Ozimek, Julio Caballero and John Williamson for valuable discussions at the earlier stages of this research. Ameya Pore thanks to Erasmus+international credit mobility programme and IISER Pune. Gerardo Aragon-Camarasa acknowledges financial support from the EPSRC Grant No. EP/S019472/1. We also acknowledge the support of NVIDIA Corporation for the donation of the Titan Xp GPU used in this research.

\bibliographystyle{IEEEtran}
\bibliography{root.bib}

\end{document}